\documentclass[a4paper,conference]{IEEEtran}
\usepackage{cite}

\ifCLASSINFOpdf
\else
\fi
\usepackage{url}
\usepackage{xspace}
\usepackage{color}
\usepackage{booktabs}
\usepackage{bm}
\usepackage{adjustbox}
\usepackage{threeparttable}
\usepackage{etoolbox}
\appto\TPTnoteSettings{\footnotesize}
\usepackage[TABBOTCAP]{subfigure}
\usepackage{CJKutf8}
\usepackage[normalem]{ulem}

\newcommand{\model}{cViL}
\newcommand{\modelbase}{\textsc{Baseline}\xspace}
\newcommand{\modelone}{\textsc{\model\textsubscript{AUG}}\xspace}
\newcommand{\modeltwo}{\textsc{\model\textsubscript{KD}}\xspace}
\newcommand{\oscar}{\textsc{Oscar}\xspace}
\newcommand{\oscarplus}{\textsc{Oscar+}\xspace}

\newcommand{\oscarplusbase}{\textsc{Oscar+\textsubscript{\textsc{B}}}\xspace}
\newcommand{\cls}{\texttt{[CLS]}\xspace}

\newcommand{\mse}{\texttt{MSE}\xspace}

\newcommand{\japanesecat}{\includegraphics[height=9pt,trim={6pt 5.5pt 6pt 2pt},clip]{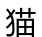}}
\newcommand{\japanesehorse}{\includegraphics[height=9pt,trim={6pt 5.5pt 6pt 2pt},clip]{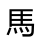}}
\newcommand{\japanesecar}{\includegraphics[height=9pt,trim={6pt 5.5pt 6pt 2pt},clip]{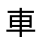}}
\newcommand{\japaneseelephant}{\includegraphics[height=9pt,trim={6pt 5.5pt 6pt 2pt},clip]{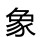}}
\newcommand{\japaneseumbrella}{\includegraphics[height=9pt,trim={6pt 5.5pt 6pt 2pt},clip]{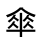}}

\newcommand{\approximately}{{\raise.17ex\hbox{$\scriptstyle\sim$}}}

\begin{document}
%

\title{\model: Cross-Lingual Training of Vision-Language Models using Knowledge Distillation}

\author{\IEEEauthorblockN{Kshitij Gupta}
\IEEEauthorblockA{IIIT Hyderabad\\
kshitij.gupta@research.iiit.ac.in}
\and
\IEEEauthorblockN{Devansh Gautam}
\IEEEauthorblockA{IIIT Hyderabad\\
devansh.gautam@research.iiit.ac.in
}
\and
\IEEEauthorblockN{Radhika Mamidi}
\IEEEauthorblockA{IIIT Hyderabad\\
radhika.mamidi@iiit.ac.in
}}


%


\maketitle

\begin{abstract}
Vision-and-language tasks are gaining popularity in the research community, but the focus is still mainly on English. We propose a pipeline that utilizes English-only vision-language models to train a monolingual model for a target language. We propose to extend \oscarplus, a model which leverages object tags as anchor points for learning image-text alignments, to train on visual question answering datasets in different languages. We
propose a novel approach to knowledge distillation to train the model in other languages \textcolor{black}{using parallel sentences}. Compared to other models that use the target language in the pretraining corpora, we can leverage an existing English model to transfer the knowledge to the target language using significantly lesser resources. \textcolor{black}{We also release a large-scale visual question answering dataset in Japanese and Hindi language.} Though we restrict our work to visual question answering, our model can be extended to any sequence-level classification task, and it can be extended to other languages as well.
This paper focuses on two languages for the visual question answering task - Japanese and Hindi.
Our pipeline outperforms the current state-of-the-art models by a relative increase of 4.4\% and 13.4\% respectively in accuracy.

\end{abstract}


\section{Introduction}
\label{sec:intro}


Several real-world problems are multimodal in nature, and vision and language are some of the most commonly researched modalities amongst them.
There have been various advancements in the vision-language domain with several techniques in V+L pretraining~\cite{DBLP:conf/nips/LuBPL19, tan2019lxmert, ChenLYK0G0020, DBLP:journals/corr/abs-1908-03557, li2019unicodervl, zhou2019unified, li2020oscar, zhang2021vinvl}, but the focus remains on English as the primary language.

The most common approach to extending these models to other languages is to train a multilingual model using several languages in the pretraining data. Although they have high cross-lingual transfer capabilities and can be finetuned to several languages, the cost of initial training of the multilingual model remains high, making it infeasible to train multilingual variants of multiple architectures. In most real-world applications, the multilingual model is used to train a monolingual model of a target language through transfer learning, and the original multilingual capabilities are no longer required.



To address these issues, we propose to directly train monolingual multimodal models in the target language with the supervision of an English model. First, we explore
simple data augmentation techniques for the task. Further, we propose a novel approach to knowledge distillation to transfer the knowledge of a high-resource English model through a machine-translated dataset. We experiment with the \oscarplus model proposed by Zhang et  al.~\cite{zhang2021vinvl} as the teacher model. We explore visual question answering in Japanese and Hindi, but our model can be extended to other sequence-level classification tasks and other languages as well. To the best of our knowledge, we are able to achieve state-of-the-art performance in both datasets. The code and dataset for our systems is available online.\footnote{\url{https://github.com/kshitij98/cViL}}

The main contributions of our work are as follows:

\begin{itemize}
    \item {We propose a novel approach to knowledge distillation by identifying the language-invariant parts of a vision-language model to train a student model in a different language. We use \oscarplus as the base model in our experiments.}

    \item We explore visual question answering in Japanese and Hindi. To the best of our knowledge, we achieve state-of-the-art performance in both the datasets with a relative increase of 4.4\% and 13.4\% respectively in accuracy.

    
    
    \item {The current state-of-the-art models use the target language in the pretraining corpora, which are very costly to train. The proposed knowledge distillation technique leverages existing English pretraining and allows to transfer the knowledge to other languages with significantly lesser resources.}

\end{itemize}



\section{Related Work}














\paragraph{Multilingual Pretraining}







{Various models~\cite{devlin2019bert}, \cite{conneau2020unsupervised} have been proposed that are pretrained on multiple languages using large multilingual corpora. They have shown state-of-the-art performance on various tasks such as cross-lingual natural language inference (XNLI) and cross-lingual Question Answering. However, these models only focus on NLP tasks and have not been designed for vision-language tasks such as Visual Question Answering and Image Text Retrieval.}

\paragraph{Vision-Language Pretraining}


A large number of multimodal pretrained models~\cite{DBLP:conf/nips/LuBPL19, li2019unicodervl, ChenLYK0G0020, zhou2019unified, li2020oscar} have been developed recently which use the transformer architectures as the backbone. However, the focus of these models remains on English only, and use monolingual datasets for training, such as MSCOCO~\cite{lin2015microsoft}, Visual Genome~\cite{krishnavisualgenome}, Conceptual Captions~\cite{sharma-etal-2018-conceptual}, and SBU Captions~\cite{DBLP:conf/nips/OrdonezKB11}.

\paragraph{Multimodal Multilingual Learning}

Multimodal and multilingual domain primarily focuses on tasks like multimodal machine translation~\cite{DBLP:journals/corr/abs-1807-11605, yao-wan-2020-multimodal, gupta-etal-2021-vita}, cross-modal retrieval, or visual question answering~\cite{DBLP:journals/corr/abs-2202-07630}.
A few recent studies~\cite{Ni_2021_CVPR, zhou2021uc} have attempted to pretrain multilingual multimodal models, which can be further finetuned on downstream tasks. However, the literature in this domain is scarce, and the current pretraining techniques to develop the multilingual variants are costly. At the same time, most of the multilingual models have been pre-trained on a limited set of languages~\cite{zhou2021uc, huang-etal-2021-multilingual}.



\paragraph{Knowledge Distillation}
{It is a widely popular technique in deep learning which is commonly used for model compression and acceleration~\cite{DBLP:journals/ijcv/GouYMT21}. Several works~\cite{DBLP:conf/aaai/FuZYTLLL21, sun2019patient} also apply knowledge distillation techniques to compress complex language models like BERT. Recently, Reimers and Gurevych~\cite{reimers-gurevych-2020-making} proposed that simple knowledge distillation can be used to transfer sentence-level embeddings to other languages as well.}



\paragraph{Visual Question Answering} It is the task of answering a given textual question about an image. It is a challenging task as it requires an understanding of both the question text and the image. VQA has been studied in various tasks such as VQA dataset~\cite{vqa_antol,balanced_vqa_v2}, Visual Genome~\cite{krishnavisualgenome}, CLEVR~\cite{DBLP:conf/cvpr/JohnsonHMFZG17} and FVQA~\cite{DBLP:journals/pami/WangWSDH18}. Although VQA has been studied extensively in English, there has been little work on VQA for other languages. Gao et al.~\cite{NIPS2015_fb508ef0} and Shimizu, Rong, and Miyazaki~\cite{C18-1163} release VQA datasets in Chinese and Japanese respectively. Gupta et al.~\cite{gupta-etal-2020-unified} create synthetic Hindi and English-Hindi code-mixed VQA datasets.




\section{Dataset Overview}
\label{sec:dataset}

The popular VQA v1.0 dataset released by Antol et al.~\cite{vqa_antol}, proposed 614k image-question pairs in the English language. Later, Goyal et al.~\cite{balanced_vqa_v2} extended it to VQA v2.0 by collecting complementary images such that every question in the dataset is associated with not just a single image, but rather a pair of similar images that result in two different answers to the question. The collection approach helped alleviate the language biases (as text alone would not be sufficient), and expanded the dataset to 1.1M question-image pairs.

We translate the balanced VQA v2.0 dataset to Japanese and Hindi using Azure Translate\footnote{\url{https://azure.microsoft.com/services/cognitive-services/translator/}} and release it as mVQA v2.0. We use the same dataset splits as Zhang et al.~\cite{zhang2021vinvl}, with 121,287 images (647,480 QA pairs) in train set, 2,000 images (10,631 QA pairs) in validation set, 36,807 images (107,394 QA pairs) in test-dev set resulting in a 60:1:18 train:validation:test split.


\section{System Overview}

%
\subsection{Model Architecture}


In this work, we base our models on the VLP method \oscarplus proposed by Zhang et al.~\cite{zhang2021vinvl}, which is an extension of the \oscar method proposed by Zhang et al.~\cite{li2020oscar}.
The motivation behind \oscar is to learn representations at the semantic level which capture modality-invariant factors.
Building on this motivation, they propose to use object tags detected in images as anchor points to ease the learning of image-text alignments.
Zhang et al.~\cite{zhang2021vinvl} further explored the visual features used to train the model and propose a better object detection model for the system.

\begin{figure}
    \centering
    \includegraphics[width=0.45\textwidth]{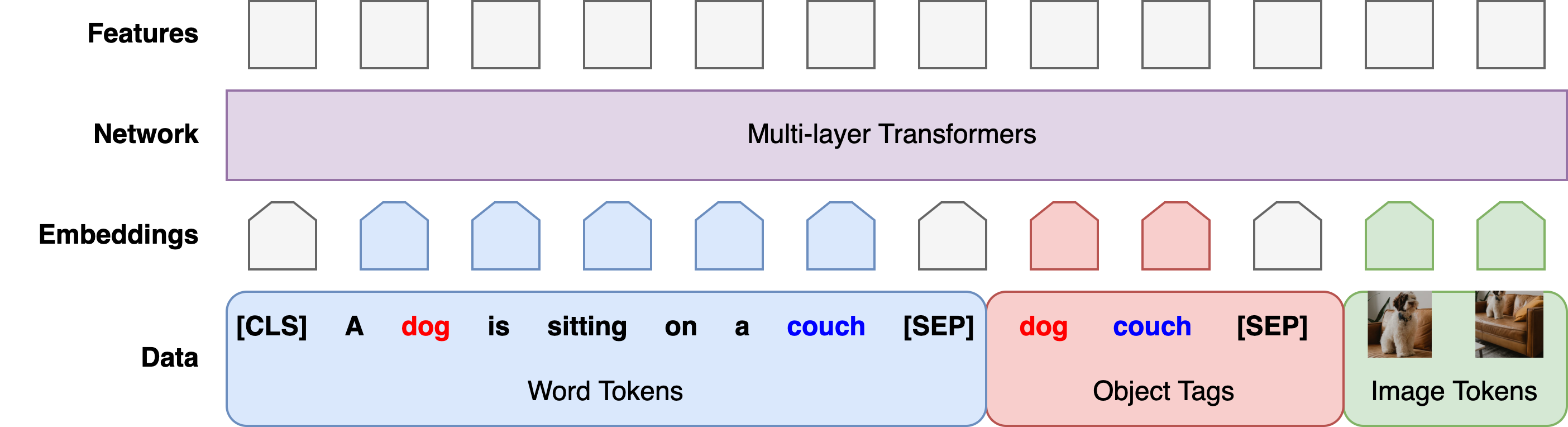}
    \caption{{A simplified representation of the model architecture of \oscar model.}}
    \label{fig:model}
\end{figure}

\oscar represents each input image-text pair as a Word-Tag-Image triple \textit{(\textbf{w}, \textbf{q}, \textbf{v})}, where \textbf{\textit{w}} is the sequence of word embeddings of the text, \textit{\textbf{q}} is the word embedding sequence of the object tags (in text) detected from the image, and \textbf{\textit{v}} is the set of region vectors of the image. 
{A simplified representation of the model architecture can be seen in Figure~\ref{fig:model}.}

The \oscarplus models have only been trained in the English language, so they cannot be used in a multilingual setting directly.
To account for non-English languages, we propose to initialize the model and the tokenizer with mBERT instead of BERT~\cite{devlin2019bert}, and then train it efficiently using knowledge distillation, instead of computationally expensive pretraining. We use the base model in our experiments with a hidden size of 768, maximum sequence length of textual tokens as 128, followed by 50 tokens for the image regions.
We use Mask R-CNN~\cite{DBLP:conf/iccv/HeGDG17} to detect the object tags, and Faster R-CNN~\cite{DBLP:conf/nips/RenHGS15} with a \texttt{X152-C4} backbone to detect the image features.

\subsection{Data Augmentation}
\label{sec:aug}

It is a widely accepted notion that larger datasets can improve the performance of deep learning models~\cite{Halevy35179,Chen8237359}. We propose augmenting the datasets for languages other than English with synthetic datasets that have been created by translating large English datasets.
\textcolor{black}{It can benefit the non-English visual question answering system by mitigating data scarcity issues and alleviating language biases (as discussed in Section~\ref{sec:dataset})}.
\textcolor{black}{Data augmentation is a crucial experiment to validate the effect of knowledge distillation, as both the methodologies will use the same set of datasets.}





\subsection{Knowledge Distillation}
\label{sec:kd}



\textcolor{black}{The motivation behind using knowledge distillation among different languages is that the image-question pair should be mapped to the same location in the semantic space irrespective of the language.}
Our main idea is to identify the language-invariant parts of a model and leverage them to train a student model for a different language by using a vision-language dataset available for both languages. {Formally, we model the method of knowledge distillation as minimizing the following objective function:}

\begin{equation}
\label{eq:kd_loss}
\mathcal{L}_{KD} =  \sum_{(s, t) \, \in \, \mathcal{X}} L\big{(}M(s), \hat{M}(t)\big{)},
\end{equation}
{where $L(\cdot)$ is a loss function that evaluates the difference between teacher and student network $M$ and $\hat{M}$, $s$ is the text input in the source language, $t$ is the corresponding input translated to the target language, and $\mathcal{X}$ denotes the parallel training dataset.} 

We propose to use the \oscarplusbase checkpoint as the teacher model in our experiments, that has been pre-trained on the medium sized corpora of \oscarplus and fine-tuned for English Visual Question Answering. The model uses VinVL visual features, the BERT tokenizer, and has a hidden size of 768.
We formulate the following types of distillations to train the student model.







\paragraph{Classification token distillation}

The \cls token captures the semantic information required for sentence-level classification tasks. Reimers and Gurevych~\cite{reimers-gurevych-2020-making} empirically study the effect of distilling \cls token embeddings to other models using a parallel dataset. The main idea is that the semantic features of the sentence are language-invariant, and we train our student model so that \cls token embeddings of the translated sentence are mapped to the same location in the vector space as the \cls token embedding of the English sentence. 
We apply \mse loss on the student's \cls token embeddings against the teacher's embeddings. {Formally, we minimize the following objective function:}


{
\begin{equation}
\label{eq:cls_loss}
\mathcal{L}_{CLS} = \texttt{MSE}(\bm{C}^{S}, \bm{C}^{T}),
\end{equation}
where $\bm{C}$ is the \cls token embedding of the teacher and student network, and {\tt MSE()} means the {\it mean squared error} loss function.
}



\paragraph{Image token distillation}

Similar to \cls token embeddings, the image token embeddings are also language-invariant and should map to the same location in the vector space in the student and teacher model.
Following this, we apply \mse loss on the corresponding 50 image token embeddings. {Formally, we minimize the following objective function:}


{
\begin{equation}
\label{eq:image_loss}
\mathcal{L}_{img} = \frac{1}{p}\sum\nolimits^{p}_{i=1} \texttt{MSE}(\bm{I}_i^{S}, \bm{I}_i^{T}),
\end{equation}
where $p$ is the number of image tokens in the input, and $\bm{I}_i$ is the $i^{th}$ image token embedding of the teacher and student network.
}





\paragraph{Object tags distillation}

The object tags are predicted in English by the same object detection model for the student and teacher model. This allows us to distill the knowledge of the corresponding embeddings, which are already aligned with the image tokens in the \textcolor{black}{feature space of the} teacher model.
Since the predicted object tags are in English, the multilinguality of mBERT can also help align the text in the target language with the image tokens by using the aligned English object tags as pivot points between the two modalities~\cite{Ni_2021_CVPR}.

Although the object tags used in the teacher and student models are identical, different subwords can be created for some tags because of different tokenizers (mBERT and BERT). We apply \mse loss only on the matching subwords of all the object tags to handle this case. {Formally, we minimize the following objective function:}


{
\begin{equation}
\label{eq:tag_loss}
\mathcal{L}_{tag} = \frac{1}{t^2}\sum\nolimits^{t}_{i=1} \sum\nolimits^{t}_{j=1} \bm{A}_{ij} \: \texttt{MSE}(\bm{O}_i^{S}, \bm{O}_j^{T}),
\end{equation}
where $t$ is the number of object tag tokens in the input, and $\bm{A}_{ij}$ is a binary matrix representing whether the object tag tokens $i$ and $j$ match and align in the student and teacher network, and $\bm{O}_i$ is the $i^{th}$ object tag token embedding.
}







\paragraph{Code-mixed distillation}

Several works~\cite{DBLP:conf/ijcai/QinN0C20, yang-etal-2020-csp, Ni_2021_CVPR, gautam-etal-2021-comet} study code-switching to augment the data for training the models in a multilingual setting. 
We propose creating code-switched translations (sentences with both English and the target language) of our data which is fed to the student model. We then apply knowledge distillation on the aligned words in the English and code-switched sentences.

{Previous works~\cite{Ni_2021_CVPR} mostly use bilingual dictionaries to translate parts of the sentence without using the rest of the sentence's context. We propose a novel approach to generate contextual translations by using a bilingual word alignment tool~\cite{dou-neubig-2021-word}}.
We perform word-level alignment between the parallel sentences and randomly replace 15\% of the words in the target language with their English alignments (we only consider the words with the same set of subwords using both the tokenizers). 
We apply \mse loss on the corresponding tokens in the code-switched sentence. {Formally, we minimize the following objective function:}


{
\begin{equation}
\label{eq:cm_loss}
\mathcal{L}_{CM} = \frac{1}{n^2}\sum\nolimits^{n}_{i=1} \sum\nolimits^{n}_{j=1} \bm{B}_{ij} \: \texttt{MSE}(\bm{H}_i^{S}, \bm{H}_j^{T}),
\end{equation}
where $n$ is the number of word token embeddings in the input, and $\bm{B}_{ij}$ is a binary matrix representing whether the word tokens $i$ and $j$ match and align in the student and teacher network, and $\bm{H}_i$ is the $i^{th}$ textual token embedding.}






\paragraph{Intermediate layers distillation}

Several works~\cite{jiao-etal-2020-tinybert, sun2019patient} show the importance of distilling intermediate layers from the teacher model to imitate the language understanding capabilities. We experiment with the same idea to transfer the knowledge to train the student model in a different language. We propose to apply distillation loss on a subset of the layers $\bm{L}$ of the model. In this work, we experiment with the combination of the third, sixth, and ninth layers along with the last layer for distillation.

{Finally, using the above distillation objectives (i.e. Equations~\ref{eq:cls_loss}, \ref{eq:image_loss}, \ref{eq:tag_loss} and \ref{eq:cm_loss}), we apply the following distillation loss on the chosen intermediate layers:}


{
\begin{equation}
\label{eq:layer_loss}
\mathcal{L}_{distil} = \sum_{m \in \bm{L}} \!\lambda_{m} ( \mathcal{L}_{CLS} + \mathcal{L}_{img} + \mathcal{L}_{tag} + \mathcal{L}_{CM} ),
\end{equation}
}

{where $\bm{L}$ is the set of layers on which the loss is applied, and $\lambda_{m}$ is the weight for the $m^{th}$ layer.}


\subsection{Training Methodology}

\begin{figure}
    \centering
    \includegraphics[width=0.85\columnwidth]{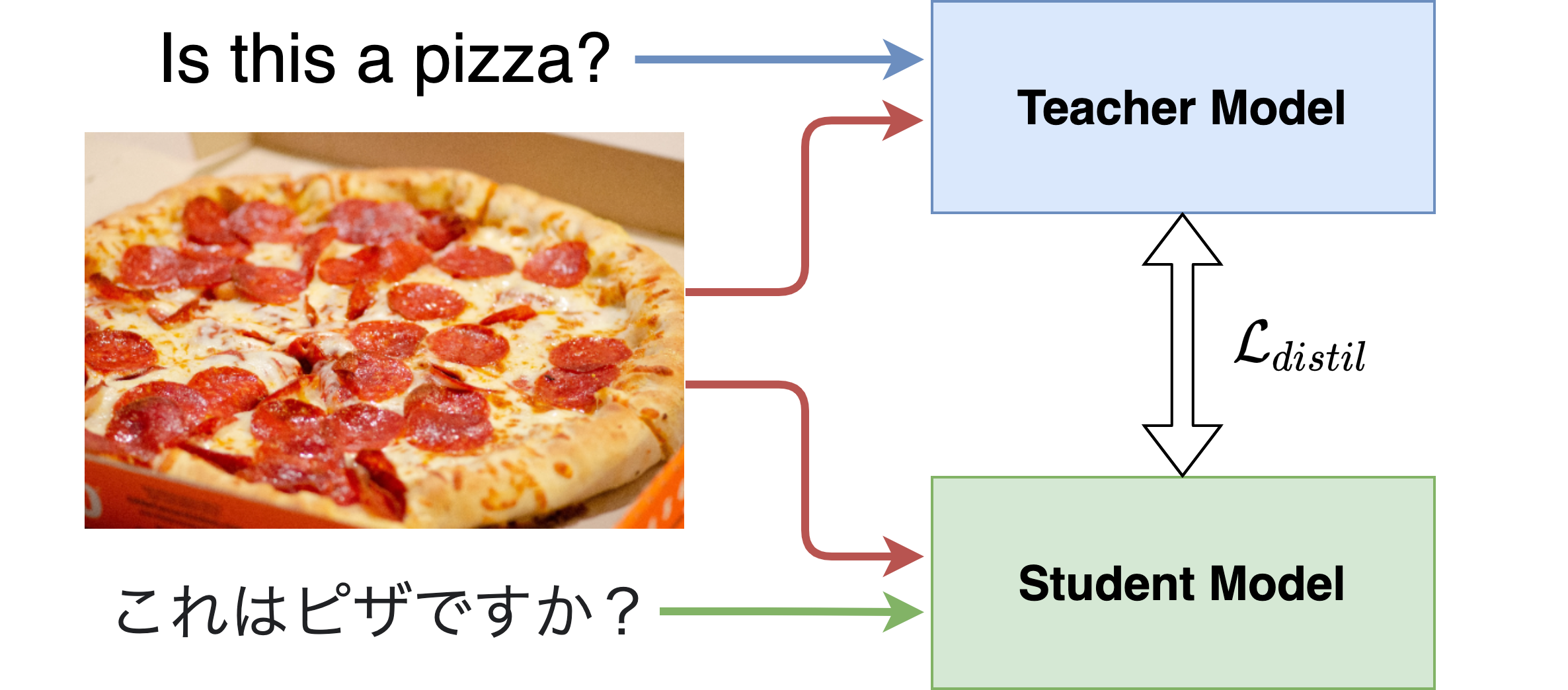}
    \caption{{We use the proposed knowledge distillation techniques to train the student in a target language.}}
    \label{fig:pipeline}
\end{figure}


In this section, we discuss different methodologies to train our model after initializing the model with mBERT. First, we propose \modelone which uses the data augmentation techniques discussed in Section~\ref{sec:aug}. We also propose \modeltwo which uses the knowledge distillation techniques discussed in Section~\ref{sec:kd}.


\paragraph{\modelone} We translate a task-specific dataset in English to the target language. Further, we propose a two-step training procedure in which we first train our model on the synthetic data created from the translation system. Finally, we finetune our model on the {real} dataset in the target language.

\paragraph{\modeltwo} We choose a task-specific dataset on which the teacher model has been trained.
Further, we propose a two-step training procedure in which we first translate that dataset to the target language to train our model using knowledge distillation  {as shown in Figure~\ref{fig:pipeline}}.
Following knowledge distillation, we finetune our model on the real dataset in the target language.

\section{Experiments}

In this section, we discuss the experiments we perform for evaluating our system. We restrict our work to Visual Question Answering, but it can be extended to any vision-language task and any language supported by mBERT. We select the VQA task for our work because it has a well-defined evaluation metric and is frequently used to test the performance of VLP models.



\subsection{Japanese Visual Question Answering\label{sec:japanesevqa}}

Shimizu, Rong, and Miyazaki~\cite{C18-1163} released the Japanese Visual Genome VQA dataset, which consists of 99,208 images from Visual Genome~\cite{krishnavisualgenome}, along with 793,664 question-answer pairs in Japanese (8 question-answer pairs per image). We split the dataset similar to Zhou et al.~\cite{zhou2021uc} with a 30:1:19 train:validation:test split consisting of 59,550 images (476,400 QA pairs) in train set, 1,984 images (15,872 QA pairs) in validation set and 37,674 images (301,392 QA pairs) in test set. We model the task as a classification problem with the 3,000 most frequent answers as the classes, which cover 74.39\% of the answers in the dataset.

We use the parallel dataset mVQA v2.0 proposed in Section~\ref{sec:dataset} for \modelone during the first step of our training. We use the 3,129 most frequent English answers of the VQA v2 as the classes. We also use this dataset in \modeltwo during the first step of our training to perform knowledge distillation. We train the student model for Japanese using the distillation objectives discussed in Section~\ref{sec:kd} from the output embeddings of the teacher model, which is fed the parallel English sentences.

We feed the output embedding of the \cls token of our models to a task-specific linear classifier. During fine-tuning, we minimize the cross-entropy loss between the predictions and the ground truth class. For evaluation, we report the accuracy of the predictions of our models. We also report the BLEU score between our predicted answers and the ground truth answers to handle the cases where there are overlapping words in the predicted answers and ground truth answers, but they do not match completely. {It can also help evaluating answers which are not covered by the set of classes of our model.}
{For a fair comparison, we use \textit{fugashi}~\cite{mccann-2020-fugashi} tokenizer and NLTK library similar to Zhou et al.~\cite{zhou2021uc} to report the BLEU score.}






\subsection{Hindi Visual Question Answering}


\textcolor{black}{There exists no human annotated dataset for visual question answering in Hindi, so we experiment with mVQA v2.0 - the translated version of VQA v2.0 dataset. As baselines, we also compare with the models trained on the translated version of VQA v1.0~\cite{gupta-etal-2020-unified}. }
We do not consider \modelone for Hindi Visual Question Answering since we are evaluating on the synthetic dataset itself. Similar to Section~\ref{sec:japanesevqa}, we use the dataset we created to train \modeltwo using knowledge distillation. We train the student model for Hindi using knowledge distillation from the output embeddings of the teacher model, which is fed the parallel English sentences.

We model the task as a multi-label classification problem. We consider the 3,129 most frequent English answers and translate these answers to Hindi. We merge the classes which have the same translation, such as \textit{up} and \textit{above} have the same Hindi translation \includegraphics[height=9pt,trim={6pt 5pt 6pt 2pt},clip]{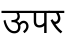} (\textit{upar}). We finally obtained 2,912 classes, which cover 92.68\% of the dataset. We would also like to handle cases where the same English word can have multiple Hindi translations in future work. Again, we feed the output embedding of the \cls token of our models to a task-specific linear classifier. During fine-tuning, we minimize the cross-entropy loss between the predictions and the ground truth classes. During the evaluation, we predict the class with the highest score and report the accuracy of the predictions of our models.

\subsection{Experimental Setup}
\label{sec:setup}

We train all our systems on 4 Nvidia GeForce RTX 2080 Ti GPUs with a batch size of 32 question-answer pairs per GPU. We use the AdamW optimizer~($\epsilon = 1e^{-8}, \beta_{1} = 0.9, \beta_{2} = 0.999, \mathrm{wd} = 0.05$)~\cite{loshchilov2018decoupled} available in PyTorch~\cite{NEURIPS2019_9015} with 0.3 dropout and 0.1 attention dropout. We lower case the text before feeding it to our models. To train our models efficiently, we use mixed-precision training (with fp16). The training methodology for both our systems is shown below:

 \paragraph{\modelone} First, we train our model on the augmented data for 5 epochs with a learning rate of $1e^{-4}$, while keeping all the layers except the final classifier layer frozen. Next, we train the whole model on the augmented dataset for 25 epochs with a learning rate of $5e^{-5}$. Then, we perform task-specific fine-tuning of the model on the real dataset for 5 epochs with a learning rate of $1e^{-4}$, while keeping all the layers except the final classifier layer frozen. Finally, we fine-tune the whole model for 15 epochs with a learning rate of $5e^{-5}$.

\paragraph{\modeltwo} First, we train using knowledge distillation for 10 epochs {(\approximately50k steps)} with a learning rate of $1e^{-4}$, and $\lambda_{m}=1$. Finally, similar to \modelone, we perform task-specific fine-tuning on the real dataset for 5 epochs with a learning rate of $1e^{-4}$, while keeping all the layers except the final classifier layer frozen, followed by fine-tuning of the whole model for 15 epochs with a learning rate of $5e^{-5}$.

We validate the models every 500 steps and select the best checkpoint based on the highest validation accuracy.


\section{Results and Discussion}

\subsection{Main Results}


We compare the performance of our systems against the current state-of-the-art models. We observe that our model outperforms the strong translate-test baseline UNITER\textsubscript{CC}~\cite{ChenLYK0G0020,zhou2021uc}, and the pretrained multilingual multimodal model UC\textsuperscript{2}~\cite{zhou2021uc}. We also compare the performance of our model against a \modelbase, in which we fine-tune our model after initializing with mBERT on the task dataset without any knowledge distillation or augmentation.

\subsubsection{Japanese Visual Question Answering}

We show the performance of our models on the Japanese Visual Genome VQA dataset in Table~\ref{tab:jap_results}. Both our systems perform better than the current state-of-the-art models. We also see that \modeltwo performs better than \modelone.
{Since the dataset coverage of the candidate classes is just 74.39\%, the final accuracy values of the classification models are heavily impacted. Even though our model classifies 48.02\% of the times correctly in the covered dataset, the final accuracy score is reduced to 35.72\%.}

\textcolor{black}{We also observe that the difference between \modelone and \modeltwo does not seem significant. Zhang et al.~\cite{zhang2021vinvl} showed an ablation of OSCAR+ without and with VLP: the scores were 71.4\% and 74.9\%. In a similar fashion, we also compared the performance of \modelone and \modeltwo on the augmented Japanese VQA v2 dataset: the scores are 69.3\% and 72\%. We see that knowledge distillation leads to an increase of 2.7\% similar to pretraining leading to an increase of 3.5\%. The differences are less prominent in the real Japanese dataset as the dataset coverage is low. Further, Reiter~\cite{reiter-2018-structured} discusses that BLEU is not a good metric to compare non machine translation tasks and models with close performance.}

\begin{table}[ht]

  \centering
  \subtable[Performance of our systems on the test set of the Japanese Visual Genome VQA dataset.\label{tab:jap_results}]{

\begin{adjustbox}{max width=\columnwidth}
\begin{threeparttable}
\begin{tabular}{lcc}
\toprule
\textbf{Model} & \textbf{Accuracy} & \textbf{BLEU} \\
\midrule
PCATT\textsuperscript{*}~\cite{C18-1163} & $19.2$ & - \\
UNITER\textsubscript{CC}~\cite{ChenLYK0G0020,zhou2021uc} & $22.7$ & $11.8$ \\
UC\textsuperscript{2}~\cite{zhou2021uc} & $34.2$ & $26.8$ \\
\midrule
\modelbase & $33.75$ & $32.5$ \\
\modelone & $34.93$ & $\textbf{33.6}$ \\
\modeltwo & $\textbf{35.72}$ & $33.1$ \\
\bottomrule
\end{tabular}
\begin{tablenotes}
\item [*] They use different train / dev / test splits.
\item [] 
\end{tablenotes}
\end{threeparttable}
\end{adjustbox}



  }
  \centering
  \subtable[Class-wise scores for the different question types in the test set of Japanese dataset. We identify the question type by checking the presence of each question word.\label{tab:classes_jap}]{

\begin{tabular}{lc}
\toprule
\textbf{Question type} & \textbf{Accuracy} \\
\midrule
\textit{nani} (what) & $38.21$ \\
\textit{dare} (who) & $35.40$ \\
\textit{doko} (where) & $23.74$ \\
\textit{donna} (what kind) & $36.10$ \\
\textit{dorekurai} (how much) & $33.17$ \\
\textit{dou} (how) & $32.24$ \\
\textit{itsu} (when) & $63.25$ \\
\textit{ikutsu} (how many) & $29.22$ \\
\textit{naze} (why) & $11.44$ \\
\bottomrule \\
\end{tabular}

  }
\caption{Performance of our systems on the Japanese Visual Genome VQA dataset.}
\end{table}



We also show the performance of our system for different question types in Table~\ref{tab:classes_jap}. We can see that our systems struggle with certain question types such as Number/ikutsu (how many), doko (where) and naze (why).







\begin{figure*}[ht]

\begin{minipage}{0.33\textwidth}
  \centering
  \subfigure[Output embeddings of the textual tokens by teacher and student model.\label{fig:tsne-text}]{
    \includegraphics[width=0.9\textwidth]{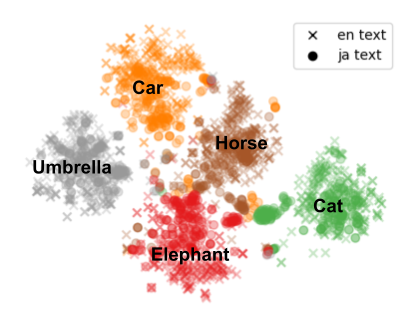}
  }
\end{minipage} 
\begin{minipage}{0.33\textwidth}
  \centering
  \subfigure[Output embeddings of the image tokens by teacher and student model.\label{fig:tsne-image}]{
    \includegraphics[width=0.9\textwidth]{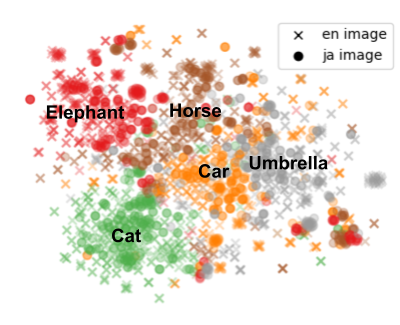}
  }
\end{minipage} 
\begin{minipage}{0.33\textwidth}
  \centering
  \subfigure[Alignment between image and textual embeddings output by our student model.\label{fig:tsne-ja}]{
    \includegraphics[width=0.9\textwidth]{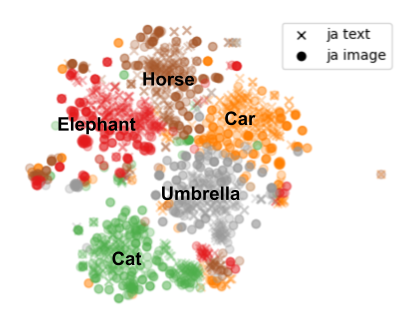}  
  }
\end{minipage}%
\caption{2D visualization of the output embeddings of the teacher and student model using t-SNE. The points from the same object class share the same color.}
\label{fig:tsnes}
\end{figure*}

\subsubsection{Hindi Visual Question Answering}

We show the performance of our models on Hindi Visual Question Answering in Table~\ref{tab:hin_results}.
{We observe the significant effect of knowledge distillation by comparing \modeltwo against the \modelbase. There is a relative increase of 7.3\% in accuracy by using knowledge distillation in VQA v2.0.}
\textcolor{black}{We also compare with BUTD~\cite{DBLP:conf/cvpr/00010BT0GZ18}, Bi-linear attention~\cite{DBLP:conf/nips/KimJZ18}, and Multimodal fusion~\cite{gupta-etal-2020-unified} on the Hindi version of VQA v1.0 where the authors fuse their multilingual question encoding and image features by adopting Bilinear Attention Networks.~\cite{DBLP:conf/nips/KimJZ18}.} Our system performs significantly better than the current state-of-the-art models.

\begin{table}[ht]
  \resizebox{4.7cm}{!}{\subtable[Performance of our systems on the Hindi version of dev set of VQA v1.0 and test-dev set of VQA v2.0.\label{tab:hin_results}]{

\begin{threeparttable}
\begin{tabular}{lc}
\toprule
\textbf{Model} & \textbf{Accuracy} \\
\midrule
\multicolumn{2}{c}{\textit{VQA v1.0}} \\ 
BUTD~\cite{DBLP:conf/cvpr/00010BT0GZ18} & $60.15$ \\
Bi-linear Attention~\cite{DBLP:conf/nips/KimJZ18} & $62.50$ \\
Multimodal Fusion~\cite{gupta-etal-2020-unified} & $64.51$ \\
\modelbase & $68.39$ \\
\modeltwo & $\textbf{73.13}$ \\
\midrule
\multicolumn{2}{c}{\textit{VQA v2.0}} \\
\modelbase & $67.71$ \\
\modeltwo & $\textbf{72.65}$ \\
\bottomrule
\end{tabular}
\begin{tablenotes}

\item [] 
\end{tablenotes}
\end{threeparttable}


  }} \resizebox{3.7cm}{!}{\subtable[Performance of our systems on the various question types in the Hindi dataset.\label{tab:classes_hin}]{

\begin{tabular}{lc}
\toprule
\textbf{Question type} & \textbf{Accuracy} \\
\midrule
Yes / No & $86.99$ \\
Number & $58.18$ \\
Other & $63.65$ \\
Overall & $72.65$ \\
\bottomrule \\

\end{tabular}

  }}
\caption{Performance of our systems on the Hindi version of VQA dataset.}
\end{table}



We also show the performance of our system for different question types in Table~\ref{tab:classes_hin}. We can see that our system performs well on Yes/No type of questions, while struggles with numerical answers.






{\paragraph{Training Cost} We are able to train our models \textcolor{black}{from scratch} using knowledge distillation with the configuration mentioned in Section~\ref{sec:setup} in just 17 hours.
The current standard approach to train models in a target language relies on a pretraining corpus containing the same language. The pretraining step is very costly and requires a high set of resources to train.
Note that English pretraining is still necessary for distillation. Our methodology can benefit in transferring the knowledge of existing English pretrained models without an extra step of pretraining for the target language.
}


\subsection{Qualitative Analysis}

We visualize the learned features for the image-question pairs using t-SNE~\cite{vandermaaten08a} in Figure~\ref{fig:tsnes}. For each image region and word token, we pass them through the model and use the last layer embeddings for each token. We compare the embeddings of \oscarplus (using the publicly available checkpoint which has been fine-tuned on VQA v2) on the VQA v2 dataset and the embeddings of \modeltwo (which has been fine-tuned on Japanese Visual Genome VQA dataset) on the Japanese Visual Genome VQA dataset. We focus on the embeddings of the following five objects since they appear frequently in the dataset and have a single subword after tokenization in both English and Japanese: \texttt{car}~(\japanesecar), \texttt{cat}~(\japanesecat), \texttt{elephant}~(\japaneseelephant), \texttt{horse}~(\japanesehorse) and \texttt{umbrella}~(\japaneseumbrella).

We compare the embeddings of the image tokens and the word tokens generated by \modeltwo for the five objects and show the results in Figure~\ref{fig:tsne-ja}. We find that for each class, the word token embeddings and the image token embeddings are close, and each class can be distinguished. We compare  the embeddings generated by the teacher and student model for image tokens and textual tokens of the five objects in Figure~\ref{fig:tsne-image} and Figure~\ref{fig:tsne-text}. We find that the textual token embeddings generated by the models align well and form easily distinguishable clusters for each object. The image token embeddings also align well and form clusters, but there is some mixture between the different clusters.









\subsection{Ablation Studies}

\begin{table}[!ht]
\begin{center}
\resizebox{0.85\columnwidth}{!}{
\begin{tabular}{lcc}
\toprule
\textbf{Model} & \textbf{Accuracy} \\
\midrule
\modeltwo & $\textbf{35.72}$ \\
\modeltwo (w/o classification token distillation) & $35.58$\\
\modeltwo (w/o code-mixed distillation) & $35.57$\\
\modeltwo (w/o image tokens distillation) & $35.38$ \\
\modeltwo (w/o object tags distillation) & $35.19$ \\
\modeltwo (w/o intermediate layers distillation) & $34.55$ \\
\modelbase & $33.75$\\
\bottomrule
\end{tabular}
}%
\end{center}
\caption{Ablation studies of the various distillation objectives.}
\label{tab:ablations}
\end{table}

We conduct ablation studies to investigate the contributions of the different distillation objectives we consider. We remove each distillation objective and compare the performance of our systems on the test set of the Japanese Visual Genome VQA dataset. We show the performance of the systems in Table~\ref{tab:ablations}. We find that each distillation is important for the performance of our system. \textcolor{black}{Intermediate layers distillation is the most crucial for the performance of our system, while object tag distillation also plays an important role which was hypothesized to help align the two modalities.}





\section{Conclusion}


In this paper, we explored {simple data augmentation techniques and proposed a novel approach to knowledge distillation to train a transformer-based VL fusion model \oscarplus in different languages. 
We are able to leverage an existing English pretrained model and transfer the knowledge to other languages with significantly lesser resources than pretraining on the target language as well.}
We explored the visual question answering task in two languages - Japanese and Hindi. To the best of our knowledge, we are able to achieve state-of-the-part performance in the VQA datasets of both languages.
We also release a large-scale visual question answering dataset in Japanese and Hindi language.
Further, we have also presented a thorough analysis of the trained models quantitatively and qualitatively. We visualise the learned embeddings of the model and later conduct ablations studies to understand the effect of each contribution.
Though we currently restrict our work to visual question answering, the model can be extended to other sequence-level classification tasks as well.

As a part of future work, we would like to explore the methodology on other models and tasks. We would also like to explore the knowledge distillation techniques more deeply and better gauge the potential of the proposed methodology.

%
\IEEEpeerreviewmaketitle

\bibliographystyle{IEEEtran}
\bibliography{IEEEabrv,anthology,custom}
%






\newpage
\appendix
\section*{Effect of Knowledge Distillation}

The motivation behind using knowledge distillation is to pass the representations learned by the English model in the resource-expensive pretraining step to the student model in another language by using lesser resources. We aim to simulate the effects of pretraining on the English model on our student models. To give context, Zhang et al.~\cite{zhang2021vinvl} report the effect of pretraining in the English teacher model, OSCAR+, and see that VLP (vision-language pretraining) only increases the accuracy to 74.9\%, while the model when fine-tuned from scratch (without VLP) on the VQA dataset benchmarks at 71.4\% accuracy. Similarly, we also report the effect of knowledge distillation on the Japanese model. We see that adding the step of knowledge distillation increases the accuracy from 69.3\% to 72\%, which resembles the pretraining trend.
To clarify, we report the performance on test-dev sets of the VQA v2.0 dataset in Table~\ref{tab:kd-effect}.

\begin{table}[ht]


\begin{adjustbox}{max width=\columnwidth}
\begin{threeparttable}
\begin{tabular}{lcccc}
\toprule
 & \textbf{VQA v2.0 EN} & \textbf{VQA v2.0 JA} & \multicolumn{2}{c}{\textbf{VG VQA JA}} \\

\textbf{Model} & \textbf{Accuracy} & \textbf{Accuracy} & \textbf{Accuracy} & \textbf{BLEU} \\
\midrule
VinVL (no VLP)~\cite{zhang2021vinvl} & $71.4$ & - & - & - \\
VinVL (\oscarplusbase)~\cite{zhang2021vinvl} & $\textbf{74.9}$ & - & - & - \\
\midrule
\modelone & - & $69.3$ & $34.93$ & $\textbf{33.6}$ \\
\modeltwo & - & $\textbf{72}$ & $\textbf{35.72}$ & $33.1$ \\
\bottomrule \\

\end{tabular}
\end{threeparttable}
\end{adjustbox}

\caption{Comparing the effect of pre-training in the teacher model and knowledge distillation in the student model}
\label{tab:kd-effect}
\end{table}

The results of the data augmentation strategy and knowledge distillation seem similar, but while modeling the visual question answering task as a classification problem, larger differences in performance are reflected as subtle differences after a certain threshold in the final accuracy because of the large number of classes (3129 classes). Further, the real Japanese dataset has a very low dataset coverage which further conceals the effect (more details in Section VI-A-1).

\section*{Dataset Analysis}

We went through a random subset of our translations manually to check the quality of the dataset. We found that nearly all the translations preserved their meaning. We believe this is because the English questions in VQA are direct and short (the average question length is only 6), and the translation system is able to translate them accurately. We did not pursue a formal analysis of the dataset as it is not one of the major contributions of our paper.

We have shown the details of the datasets used for training and fine-tuning UC\textsubscript{2}, \modelone and \modeltwo in Table~\ref{tab:datadetails}. We can see that we use the same dataset splits as UC\textsuperscript{2} for fine-tuning on the Japanese Visual Genome dataset. The \oscarplusbase checkpoint we use for distilling \modeltwo has been pre-trained on a smaller corpus as compared to UC\textsuperscript{2}, and the mVQA v2.0 dataset which we use for Knowledge Distillation / Augmented training is also very small as compared to the pretraining corpus of UC\textsuperscript{2}.

\begin{table}[ht]
\begin{adjustbox}{max width=\columnwidth}
\begin{threeparttable}
\begin{tabular}{lccc}
\toprule
                                & \textbf{UC\textsuperscript{2}}    & \textbf{\modelone} & \textbf{\modeltwo} \\ \midrule
\multicolumn{4}{l}{Pretraining/KD}                                                           \\ \midrule
\textbf{Pretraining Datasets}            & 3.3M (19.8M)    & -                   & 1.89M (4.87M)      \\
\textbf{mVQA v2.0} & -               & 0.12M (0.65M)       & 0.12M (0.65M)      \\ 
\textbf{Total}                  & 3.3M (19.8M)    & 0.12M (0.65M)       & 1.93M (4.98M) \textsuperscript{*}      \\ \midrule
\multicolumn{4}{l}{Fine-tuning on Japanese Visual Genome dataset}                                    \\ \midrule
\textbf{Training Set}                  & 59k (476k)   & 59k (476k)       & 59k (476k)      \\
\textbf{Validation Set}                    & 2k (16k) & 2k (16k)     & 2k (16k)    \\
\textbf{Test Set}                   & 38k (301k)    & 38k (301k)        & 38k (301k)       \\ 
\textbf{Total}                  & 99k (793k)    & 99k (793k)        & 99k (793k)        \\ \bottomrule
\end{tabular}
\begin{tablenotes}
\item [*] There is an overlap of 83k images and 545k QA pairs in the pre-training dataset of \oscarplus and mVQA v2.0.
\item [] 
\end{tablenotes}
\end{threeparttable}
\end{adjustbox}
\caption{The values are reported in the format: \#images (\#captions and QA pairs).
\label{tab:datadetails}}
\end{table}

\end{document}